\documentclass[lettersize,journal]{IEEEtran}
\usepackage{amsmath,amsfonts}
\usepackage{algorithmic}
\usepackage{algorithm}
\usepackage{array}
\usepackage[caption=false,font=normalsize,labelfont=sf,textfont=sf]{subfig}
\usepackage{textcomp}
\usepackage{stfloats}
\usepackage{url}
\usepackage{verbatim}
\usepackage{graphicx}
\usepackage{cite}
\usepackage{setspace}
\usepackage{xcolor}
\usepackage[hidelinks]{hyperref}
\usepackage{orcidlink}
\usepackage[T1]{fontenc}
\usepackage{booktabs}   
\usepackage{siunitx}

\begin{document}
%E2VD: 
\title{Energy-Efficient Vision Transformer Inference for Edge-AI Deployment}

\author{
{Nursultan Amanzhol}\textsuperscript{1} and 
%{Fariza Temirkhanova}\IEEEauthorrefmark{1}, and
{Jurn-Gyu Park}\textsuperscript{1}\textsuperscript{*}
%\IEEEauthorrefmark{1}
%Authors \orcidlink{0009-0007-xxxx-xxxx},
%Meruyert Karzhaubayeva \orcidlink{0009-0007-4811-7418}, 
%Aidar Amangeldi \orcidlink{0009-0009-5198-4680}, and 
%Jurn-Gyu Park %\orcidlink{0000-0002-7186-1700}
% %,~\IEEEmembership{Member,~IEEE,}

        % <-this % stops a space
\thanks{\textsuperscript{1}The School of Engineering and Digital Sciences, Nazarbayev University, Astana, Kazakhstan (\textsuperscript{*}corresponding author: jurn.park@nu.edu.kz).}% <-this % stops a space
%\thanks{The authors are with the School of Engineering and Digital Sciences (SEDS),
%Nazarbayev University, Astana, Kazakhstan, E-mail:
%meruyert.karzhaubayeva@nu.edu.kz, aidar.amangeldi@nu.edu.kz, and jurn.park@nu.edu.kz (Corresponding author).}
%\thanks{Manuscript received April 19, 2021; revised August 16, 2021.}
}

% The paper headers
%\markboth{IEEE Embedded System Letters,~Vol.~XX, No.~YY, MONTH~2023}%
%{Shell \MakeLowercase{\textit{et al.}}: A Sample Article Using IEEEtran.cls for IEEE Journals}

%\IEEEpubid{0000--0000/00\$00.00~\copyright~2023 IEEE}

\maketitle
\begin{spacing}{0.9}
\begin{abstract}
The growing deployment of Vision Transformers (ViTs) on energy-constrained devices requires evaluation methods that go beyond accuracy alone. 
%E3P‑ViT
We present a two-stage pipeline for assessing ViT energy efficiency that combines device-agnostic model selection with device-related measurements. We benchmark 13 ViT models on ImageNet‑1K and CIFAR‑10, running inference on NVIDIA Jetson TX2 (edge device) and an NVIDIA RTX 3050 (mobile GPU). 
The device-agnostic stage uses the NetScore metric for screening; the device-related stage ranks models with the Sustainable Accuracy Metric (SAM). Results show that hybrid models such as LeViT\_Conv\_192 reduce energy by up to 53\% on TX2 relative to a ViT baseline (e.g., SAM5=1.44 on TX2/CIFAR‑10), while distilled models such as TinyViT‑11M\_Distilled excel on the mobile GPU (e.g., SAM5=1.72 on RTX 3050/CIFAR‑10 and SAM5=0.76 on RTX 3050/ImageNet‑1K).
\end{abstract}


\section{Introduction}

\IEEEPARstart{R}{ecently}, Vision Transformers (ViTs) have emerged as the state-of-the-art in many of computer vision tasks, from image classification to object detection \cite{dosovitskiy2020image}. However, the power of ViTs comes at a significant cost. The quadratic complexity of the self-attention mechanism with respect to input sequence length leads to substantial computational and memory requirements. This presents a challenge to their deployment on edge devices. Consequently, a lot of research is emerging towards developing efficient ViTs, such as EfficientViT \cite{cai2023efficientvit}, TinyViT \cite{wu2022tinyvit}, LeViT \cite{graham2021levit}, etc., using knowledge distillation, pruning and hybrid designs \cite{xu2021survey}.
Despite progress, evaluating energy efficiency remains difficult. Common device‑agnostic metrics—MACs, parameter count, and accuracy—are helpful for screening but do not capture interactions between model architecture and hardware (e.g., memory traffic, kernel fusion, bandwidth limits). We therefore propose the Energy Efficiency Evaluation Pipeline
for Vision Transformer (E3P‑ViT), a structured two‑stage pipeline that first narrows candidates using NetScore, then measures energy/time on target hardware and ranks models with SAM.

The contributions of the paper are as follows:

\begin{itemize}
{
\item Propose a device-agnostic selection stage within E3P‑ViT that uses NetScore to rank models before deployment.

\item Propose a device-related evaluation stage, within the E3P-ViT framework, that measures time/power/energy on hardware and ranks models using SAM.

\item Our results show improvements up to 53\% in terms of energy consumption for hybrid models such as LeViT\_Conv\_192 using our framework compared to a standard ViT. 
}
\end{itemize}

\section{Motivation and Related Work}
\subsection{Motivation}

Our two-stage framework addresses the disconnect between theoretical metrics and real-world efficiency. We illustrate this by comparing EfficientViT-B1 and LeViT\_Conv\_192 on an NVIDIA Jetson TX2 for CIFAR-10.

\begin{itemize}
\item \textbf{Device-Agnostic Evaluation (Theoretical Complexity)}. EfficientViT‑B1 requires 0.52G MACs and 9.1M parameters, lower than LeViT‑Conv‑192 (0.70G MACs, 10.9M params): 26\% fewer MACs and 16.5\% fewer parameters.
\item \textbf{Device-Related Evaluation (Empirical Measurement)}. On the device, LeViT‑Conv‑192 finishes in 295.36 ms, whereas EfficientViT‑B1 takes 324.39 ms — 9.8\% slower — thus consuming more energy for the same task.

\end{itemize}

\begin{figure}[!t]
\centering
\subfloat[Device-Agnostic Evaluation]{\includegraphics[width=0.48\columnwidth]{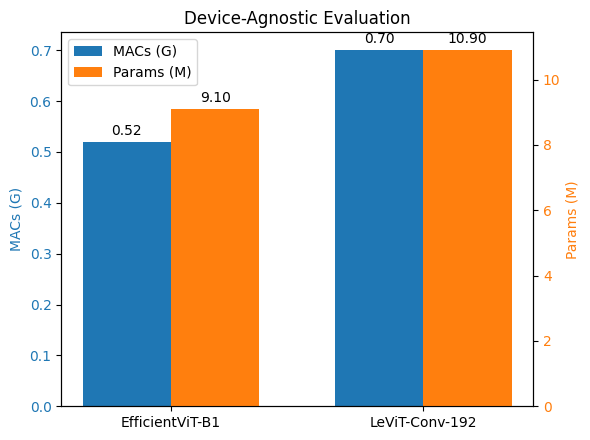}\label{motivating-example:a}}
\hfill
\subfloat[Device-Related Evaluation]{\includegraphics[width=0.48\columnwidth]{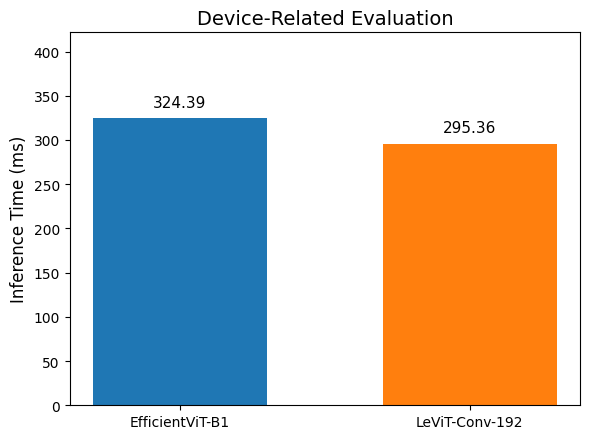}\label{motivating-example:b}}
\caption{Motivating Example: Comparison Between Device-Agnostic and Device-Related Metrics. While EfficientViT-B1 has fewer theoretical MACs and parameters (a) LeViT\_Conv\_192 achieves faster inference on hardware (b).}
\label{motivating-example}
\end{figure}

Figure~\ref{motivating-example} visualizes this gap. Energy depends not only on computational load (MACs/params) but also on measured power and latency. E3P‑ViT addresses this by coupling device-agnostic screening with device-related evaluation. Code is available here.\footnote{\url{https://github.com/nursultanamanzholdev/e2vd-energy-efficient-vit}}

\subsection{Related Work}

Prior work reduces ViT complexity through architectural changes, including knowledge distillation (DeiT, TinyViT) \cite{touvron2021training,wu2022tinyvit}, hybrid convolution-attention designs (LeViT, PoolFormer) \cite{graham2021levit,yu2022poolformerv2}, and multi-scale attention (EfficientViT) \cite{cai2023efficientvit}. However, while these strategies successfully lower MACs and parameters, theoretical reductions do not guarantee energy savings on real devices.

Complementary work focuses on deployment optimizations. Token reduction lowers memory and attention costs (DynamicViT, EViT, ToMe) \cite{rao2021dynamicvit,liang2022evit,bolya2023tome}, while \emph{quantization} reduces arithmetic intensity (PTQ4ViT, Q-ViT) \cite{yuan2022ptq4vit,li2022qvit}. Kernel/compiler co-design mitigates runtime overhead via I/O optimization and fusion (FlashAttention, TVM) \cite{dao2022flashattention,chen2018tvm}, and hardware-oriented network co-design (TRT-ViT) targets GPU latency \cite{xia2022trtvit}. These orthogonal strategies highlight why theoretical metrics like FLOPs often fail to predict on-device energy.

FLOPs and parameter counts often fail to correlate with real-world performance due to memory and bandwidth bottlenecks \cite{hooker2021moving}. This disconnect necessitates hardware-aware evaluation \cite{xu2021survey,schwartz2020greenai}, particularly as attention costs interact with memory hierarchies in ways FLOPs miss \cite{tay2020efficient}. While composite metrics like NetScore \cite{wong2018netscore} omit hardware effects, energy-integrated metrics like SAM \cite{gowda2024watt} justify a pipeline combining device-agnostic screening with empirical measurement.

Our work is different than efficient ViT design papers that primarily optimize MACs/params \cite{touvron2021training,wu2022tinyvit,graham2021levit,yu2022poolformerv2}, and composite device‑agnostic metrics like NetScore alone \cite{wong2018netscore}, as we integrate device‑agnostic selection with empirical measurements and SAM‑based ranking to provide hardware‑specific, deployment‑ready recommendations.

\section{Methodology}
Our methodology is composed of two main stages of the E3P-ViT framework, shown in Figure~\ref{methodology-overview}: 1) device-agnostic stage and 2) device-related stage.

\bgroup
%\vspace{-4mm}
\begin{figure}[!htbp] %
\centering \makeatletter\IfFileExists{images/E3P-ViT_framework5.jpg}{\includegraphics[width=1.0\columnwidth]{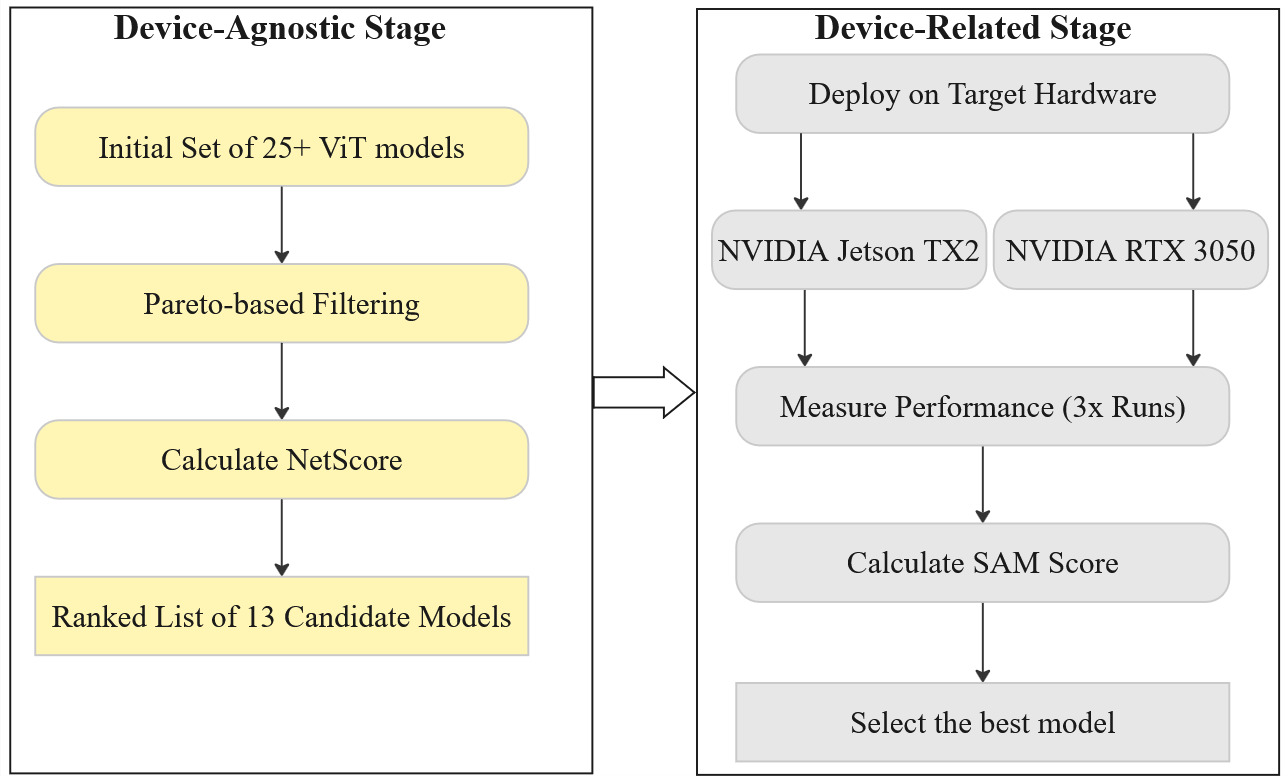}}{}
\makeatother 
\vspace{-1mm}
\caption{Methodology Overview: The Energy Efficiency Evaluation Pipeline for Vision Transformers (E3P-ViT).}
\label{methodology-overview}
\end{figure}
\egroup

\subsection{Device-Agnostic Stage}
The first stage of the E3P-ViT framework addresses the challenge of identifying suitable models from a vast and growing number of ViT architectures without requiring access to the target hardware.

\subsubsection{Framework Protocol}
\begin{figure}[!htbp]
  \centering
  \IfFileExists{images/pareto_analysis.png}{%
    \includegraphics[width=1.0\columnwidth]{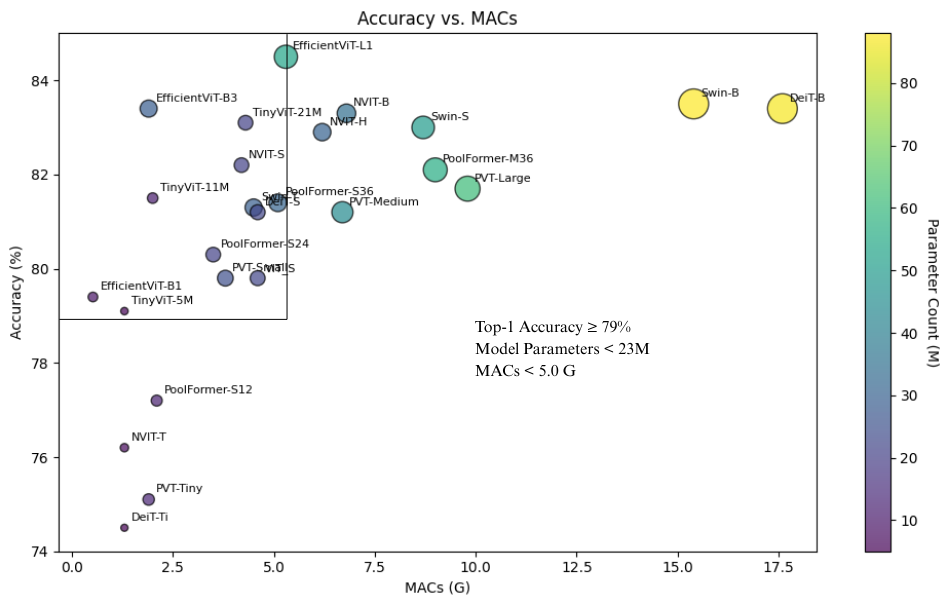}%
  }{%
    \fbox{Missing image: images/pareto_analysis.png}%
  }
  \vspace{-3mm}
  \caption{The Pareto-based filtering applied on the initial set of 25 models.}
  \label{pareto-analysis}
\end{figure}

We surveyed recent efficient ViTs and compiled 25 models spanning compact, hybrid, and distilled families, all pre‑trained on ImageNet‑1K. We then applied hard thresholds on performance and complexity to filter candidates (Pareto plot in Fig.~\ref{pareto-analysis}).

\subsubsection{Quantitative Metrics}
We selected the 13 candidate models that satisfy the thresholds of the quantitative metrics:
\begin{itemize}

\item \textbf{Top-1 Accuracy} $\geq$ 79\%: ensures a high standard of performance
\item \textbf{Model Parameters} < 23M: excludes  large models that are not suitable for deployment on edge devices
\item \textbf{MACs} < 5.0 G: constrains the computational complexity.

\end{itemize}

We rank the 13 candidates for device-related evaluation using NetScore \eqref{netscore-equation}, a device-agnostic metric that rewards accuracy while penalizing parameters and MACs.

\vspace{-1mm}
\let\saveeqnno\theequation
\let\savefrac\frac
\def\dispfrac{\displaystyle\savefrac}
\begin{eqnarray}
\let\frac\dispfrac
\gdef\theequation{1}
\let\theHequation\theequation
\label{netscore-equation}
\begin{array}{@{}l}
NetScore \;=\; 20 \log_{10}\!\left(\frac{Accuracy^{2}}
{\sqrt{Params}\;\times\;\sqrt{MACs}}\right)
\end{array}
\end{eqnarray}
\global\let\theequation\saveeqnno
\addtocounter{equation}{-1}\ignorespaces 
\vspace{-1.5mm}

\subsection{Device-Related Stage}
The second stage deploys candidate models on target hardware to measure real-world performance and energy.

\subsubsection{Framework Protocol}
Candidates are deployed on target hardware; we run inference on a representative 1000‑image subset and measure each model three times, reporting means.

\subsubsection{Quantitative Metrics}
During inference, we collect the following key device-related metrics:
\begin{itemize}
\item \textbf{Total Inference Time (s)}: The wall-clock time to complete the inference task.
\item \textbf{Average Power (mW)}: The average power drawn by the device's compute system during inference.
\item \textbf{Total Energy (J)}: Calculated as \ensuremath{Average Power} \ensuremath{\cdot} \ensuremath{Total Inference Time}.

\end{itemize}

We rank models using the Sustainable Accuracy Metric (SAM) \eqref{sam-equation}, where parameters \ensuremath{a} and \ensuremath{b} control accuracy scaling. We empirically selected \ensuremath{a=5} and \ensuremath{b=5}, which can for optimal model distinction.

\vspace{-1mm}
\let\saveeqnno\theequation
\let\savefrac\frac
\def\dispfrac{\displaystyle\savefrac}
\begin{eqnarray}
\let\frac\dispfrac
\gdef\theequation{2}
\let\theHequation\theequation
\label{sam-equation}
\begin{array}{@{}l}
SAM \;=\; \frac{b \times Accuracy^{a}}{\log_{10}(Energy)}
\end{array}
\end{eqnarray}
\global\let\theequation\saveeqnno
\addtocounter{equation}{-1}\ignorespaces 
\vspace{-2.5mm}

\section{Experimental Setup}
\textbf{Hardware:} In our study an NVIDIA Jetson TX2 (Pascal GPU, Denver/A57 CPUs, JetPack v4.5.1) is utilized. Power and time were measured using \textit{tegrastats}. Second device is an NVIDIA RTX 3050 4GB based laptop. Power and time were measured using \textit{nvidia-smi}.
\textbf{Models:} The 13 candidates from the device‑agnostic stage (Table~\ref{tab:device_agnostic}) plus a standard ViT\_S baseline.
\textbf{Fine-Tuning:}  To analyze efficiency across dataset scales, we fine-tuned the 13 ImageNet-1K pre-trained models on CIFAR-10 using two NVIDIA T4 GPUs on Kaggle.
\textbf{Datasets:} ImageNet‑1K (validation set, 1K images) and CIFAR‑10 (test set, 1K images) for inference.

\section{Results and Analysis}
We first analyze the device‑agnostic filtering stage, then the device‑related measurements on two devices and two datasets.

\subsection{Device-Agnostic Analysis}
13 candidates (Table~\ref{tab:device_agnostic}) are selected using Pareto constraints and NetScore ranking. 
EfficientViT-B1 achieved the highest NetScore (69.24) due to minimal MACs (0.52G). TinyViT-5M models ranked second due to the lowest parameter count (5.4M). In contrast, TinyViT-21M\_Distilled attained peak accuracy (84.80\%) but ranked lower due to complexity. LeViT models also ranked highly, balancing low MACs (0.7G) with moderate accuracy.

\textbf{Takeaway:} The device-agnostic stage reduced 25 ViT variants to 13 candidates via constraints and NetScore ranking, collecting distilled (DeiT, TinyViT), hybrid (LeViT, PoolFormerV2), and compact (EfficientViT) architectures for hardware evaluation.

\begin{table}[!t]
\caption{Device-agnostic ranking of candidate ViT models by NetScore (ImageNet-1K). Sorted by \textbf{NetScore}.}
\label{tab:device_agnostic}
\centering
\scriptsize
\setlength{\tabcolsep}{2pt}
\renewcommand{\arraystretch}{0.95}
\resizebox{\columnwidth}{!}{%
\begin{tabular}{@{} l
S[table-format=2.1,detect-weight=true,detect-mode=true]
S[table-format=1.2,detect-weight=true,detect-mode=true]
S[table-format=2.2,detect-weight=true,detect-mode=true]
S[table-format=2.2,detect-weight=true,detect-mode=true]
@{}}
\toprule
\textbf{Model} & {\textbf{Params (M) $\downarrow$}} & {\textbf{MACs (G) $\downarrow$}} & {\textbf{Acc (\%) $\uparrow$}} & {\textbf{NetScore $\uparrow$}} \\
\midrule
EfficientViT-B1            & 9.1  & \bfseries 0.52 & 79.40 & \bfseries 69.24 \\
TinyViT-5M\_Distilled      & \bfseries 5.4 & 1.30 & 80.70 & 67.81 \\
TinyViT-5M                 & \bfseries 5.4 & 1.30 & 79.10 & 67.46 \\
LeViT\_192                 & 10.9 & 0.70 & 79.86 & 67.27 \\
LeViT\_Conv\_192           & 10.9 & 0.70 & 79.86 & 67.27 \\
TinyViT-11M\_Distilled     & 11.0 & 2.00 & 83.20 & 63.38 \\
TinyViT-11M                & 11.0 & 2.00 & 81.50 & 63.02 \\
PoolFormerV2-S24           & 21.3 & 3.40 & 80.70 & 57.68 \\
TinyViT-21M\_Distilled     & 21.2 & 4.30 & \bfseries 84.80 & 57.54 \\
TinyViT-21M                & 21.2 & 4.30 & 83.10 & 57.19 \\
DeiT-Small\_Distilled      & 22.4 & 4.60 & 81.20 & 56.25 \\
ViT\_S (Baseline)          & 22.0 & 4.60 & 79.80 & 56.03 \\
DeiT-Small                 & 22.4 & 4.60 & 79.80 & 55.95 \\
\bottomrule
\end{tabular}}
\end{table}

\begin{table}[!t]
\caption{Device-related performance on Jetson TX2 for ImageNet-1K and CIFAR-10. Models are ranked by \textbf{SAM5}. SAM1 uses $a=b=1$; SAM5 uses $a=b=5$.}
\label{tab:device_related_tx2_both}
\centering
\scriptsize
\setlength{\tabcolsep}{2pt}
\renewcommand{\arraystretch}{0.95}
\resizebox{\columnwidth}{!}{%
\begin{tabular}{@{} l
S[table-format=2.2,detect-weight=true,detect-mode=true]  % Acc
S[table-format=4.3,detect-weight=true,detect-mode=true]  % T (s)
S[table-format=4.2,detect-weight=true,detect-mode=true]  % P (mW)
S[table-format=4.2,detect-weight=true,detect-mode=true]  % E (J)
S[table-format=1.2,detect-weight=true,detect-mode=true]  % SAM5
S[table-format=1.2,detect-weight=true,detect-mode=true]  % SAM1
@{}}
\toprule
\multicolumn{6}{c}{\textbf{ImageNet-1K}} \\
\midrule
\textbf{Model} & \textbf{Acc (\%)} & \textbf{Time (s)} & \textbf{Power (mW)} & \textbf{Energy (J)} & \textbf{SAM5} & \textbf{SAM1} \\
\midrule
TinyViT-11M\_Distilled & 83.80 & 1008.335 & 2465.35 & 2485.81 & \bfseries 0.61 & 0.25 \\
DeiT-Small\_Distilled  & 82.40 & 673.523  & 3666.47 & 2469.56 & 0.56 & 0.24 \\
TinyViT-21M            & 83.00 & 1192.005 & 3345.32 & 3987.62 & 0.55 & 0.23 \\
TinyViT-21M\_Distilled & 83.00 & 1195.911 & 3339.37 & 3993.53 & 0.55 & 0.23 \\
PoolFormerV2-S24       & 80.80 & 962.028  & 1797.96 & 1729.03 & 0.53 & 0.25 \\
TinyViT-11M            & 81.00 & 1012.115 & 2468.11 & 2497.97 & 0.51 & 0.24 \\
ViT\_S (Baseline)      & 80.90 & 733.977  & 3337.02 & 2449.30 & 0.51 & 0.24 \\
LeViT\_192             & 79.10 & 605.291  & 1763.96 & \bfseries 1067.83 & 0.51 & \bfseries 0.26 \\
LeViT\_Conv\_192       & 79.10 & \bfseries 500.187 & 2138.08 & 1069.49 & 0.51 & \bfseries 0.26 \\
DeiT-Small             & 80.80 & 667.334  & 3644.68 & 2432.21 & 0.51 & 0.24 \\
EfficientViT-B1        & 79.10 & 848.719  & \bfseries 1391.42 & 1180.22 & 0.50 & 0.26 \\
TinyViT-5M\_Distilled  & 80.10 & 825.348  & 2438.38 & 2012.50 & 0.50 & 0.24 \\
TinyViT-5M             & 78.90 & 828.584  & 2430.16 & 2013.60 & 0.46 & 0.24 \\
\midrule
\multicolumn{6}{c}{\textbf{CIFAR-10}} \\
\midrule
\textbf{Model} & \textbf{Acc (\%)} & \textbf{Time (s)} & \textbf{Power (mW)} & \textbf{Energy (J)} & \textbf{SAM5} & \textbf{SAM1} \\
\midrule
LeViT\_Conv\_192       & 97.10 & \bfseries 295.355 & 3389.26 & \bfseries 1001.04 & \bfseries 1.44 & \bfseries 0.32 \\
ViT\_S (Baseline)      & 98.60 & 452.256 & 4730.66 & 2139.47 & 1.40 & 0.30 \\
LeViT\_192             & 96.30 & 315.243 & 3350.08 & 1056.09 & 1.37 & 0.32 \\
TinyViT-5M\_Distilled  & 98.00 & 582.539 & 3645.22 & 2123.48 & 1.36 & 0.29 \\
TinyViT-11M\_Distilled & 98.60 & 736.622 & 3745.02 & 2758.66 & 1.35 & 0.29 \\
DeiT-Small             & 98.30 & 623.113 & 4042.10 & 2518.69 & 1.35 & 0.29 \\
PoolFormerV2-S24       & 97.50 & 474.328 & 4020.82 & 1907.18 & 1.34 & 0.30 \\
DeiT-Small\_Distilled  & 98.10 & 620.290 & 4017.73 & 2492.16 & 1.34 & 0.29 \\
TinyViT-5M             & 97.60 & 580.317 & 3636.69 & 2110.43 & 1.33 & 0.29 \\
TinyViT-21M\_Distilled & \bfseries 99.30 & 1070.097 & 3934.05 & 4209.82 & 1.33 & 0.27 \\
TinyViT-11M            & 98.10 & 739.820 & 3721.27 & 2753.07 & 1.32 & 0.29 \\
TinyViT-21M            & 98.30 & 1070.172 & 3921.38 & 4196.55 & 1.27 & 0.27 \\
EfficientViT-B1        & 94.80 & 324.389 & \bfseries 3338.04 & 1082.82 & 1.26 & 0.31 \\
\bottomrule
\end{tabular}}
\end{table}

\begin{table}[!t]
\caption{Device-related performance on NVIDIA RTX 3050 for ImageNet-1K and CIFAR-10. Models are ranked by \textbf{SAM5}. SAM1 uses $a=b=1$; SAM5 uses $a=b=5$.}
\label{tab:device_related_rtx_both}
\centering
\scriptsize
\setlength{\tabcolsep}{2pt}
\renewcommand{\arraystretch}{0.95}
\resizebox{\columnwidth}{!}{%
\begin{tabular}{@{} l
S[table-format=2.2,detect-weight=true,detect-mode=true]  % Acc
S[table-format=2.3,detect-weight=true,detect-mode=true]  % T (s)
S[table-format=5.2,detect-weight=true,detect-mode=true]  % P (mW)
S[table-format=3.2,detect-weight=true,detect-mode=true]  % E (J)
S[table-format=1.2,detect-weight=true,detect-mode=true]  % SAM5
S[table-format=1.2,detect-weight=true,detect-mode=true]  % SAM1
@{}}
\toprule
\multicolumn{6}{c}{\textbf{ImageNet-1K}} \\
\midrule
\textbf{Model} & \textbf{Acc (\%)} & \textbf{Time (s)} & \textbf{Power (mW)} & \textbf{Energy (J)} & \textbf{SAM5} & \textbf{SAM1} \\
\midrule
TinyViT-11M\_Distilled & 83.80 & 25.983 & 19745.50 & 512.96 & \bfseries 0.76 & \bfseries 0.31 \\
TinyViT-21M\_Distilled & 83.00 & 26.906 & 25558.65 & 687.72 & 0.69 & 0.29 \\
TinyViT-21M            & 83.00 & 26.500 & 26142.77 & 692.78 & 0.69 & 0.29 \\
DeiT-Small\_Distilled  & 82.40 & 21.338 & 28727.38 & 612.70 & 0.68 & 0.30 \\
TinyViT-11M            & 81.00 & 26.542 & 19324.59 & 512.89 & 0.64 & 0.30 \\
DeiT-Small             & 80.80 & \bfseries 21.171 & 28334.00 & 599.81 & 0.62 & 0.29 \\
TinyViT-5M\_Distilled  & 80.10 & 25.946 & 18267.54 & 473.85 & 0.62 & 0.30 \\
ViT\_S (Baseline)       & 80.90 & 25.057 & 26938.66 & 673.76 & 0.61 & 0.29 \\
PoolFormerV2-S24        & 80.80 & 33.705 & 20010.15 & 674.43 & 0.61 & 0.29 \\
LeViT\_Conv\_192        & 79.10 & 28.248 & \bfseries 14395.18 & \bfseries 406.66 & 0.59 & 0.30 \\
LeViT\_192              & 79.10 & 31.403 & 14673.82 & 460.88 & 0.58 & 0.30 \\
EfficientViT-B1         & 79.10 & 30.394 & 15720.70 & 477.76 & 0.58 & 0.30 \\
TinyViT-5M              & 78.90 & 25.744 & 19411.08 & 499.89 & 0.57 & 0.29 \\
\midrule
\multicolumn{6}{c}{\textbf{CIFAR-10}} \\
\midrule
\textbf{Model} & \textbf{Acc (\%)} &\textbf{Time (s)} & \textbf{Power (mW)} & \textbf{Energy (J)} & \textbf{SAM5} & \textbf{SAM1} \\
\midrule
TinyViT-11M\_Distilled & 98.60 & 23.779 & 21408.38 & 509.07 & \bfseries 1.72 & 0.36 \\
TinyViT-5M\_Distilled  & 98.00 & 23.417 & 19921.62 & 466.51 & 1.69 & 0.37 \\
TinyViT-11M            & 98.10 & 23.896 & 21240.06 & 507.55 & 1.68 & 0.36 \\
TinyViT-5M             & 97.60 & 23.601 & 19967.09 & 471.24 & 1.66 & 0.37 \\
ViT\_S (Baseline)      & 98.60 & 25.196 & 26304.18 & 662.75 & 1.65 & 0.35 \\
TinyViT-21M\_Distilled & \bfseries 99.30 & 25.379 & 33020.40 & 838.01 & 1.65 & 0.34 \\
LeViT\_Conv\_192       & 97.10 & 25.893 & 16049.40 & \bfseries 415.57 & 1.65 & \bfseries 0.37 \\
DeiT-Small             & 98.30 & \bfseries 19.871 & 37394.76 & 743.06 & 1.60 & 0.34 \\
TinyViT-21M            & 98.30 & 24.743 & 31781.92 & 786.38 & 1.58 & 0.34 \\
DeiT-Small\_Distilled  & 98.10 & 20.424 & 37475.37 & 765.40 & 1.57 & 0.34 \\
LeViT\_192             & 96.30 & 28.574 & \bfseries 15519.07 & 443.43 & 1.56 & 0.36 \\
PoolFormerV2-S24       & 97.50 & 33.189 & 21982.71 & 729.59 & 1.54 & 0.34 \\
EfficientViT-B1        & 94.80 & 26.270 & 17032.04 & 447.44 & 1.44 & 0.36 \\
\bottomrule
\end{tabular}}
\end{table}

\subsection{Device-Related Analysis}

13 candidates are deployed and evaluated on two devices and datasets, reporting SAM5 (\(a=b=5\)) and SAM1 (\(a=b=1\)) to contrast efficiency when accuracy is prioritized vs. deemphasized. A key finding is that the device-agnostic leader, EfficientViT-B1, is not the top performer in any scenario. Instead, models with lower NetScores, like TinyViT-11M\_Distilled and LeViT\_Conv\_192, consistently rank at the top. This empirically validates the device–agnostic gap and confirms the necessity of device-related measurements \cite{hooker2021moving,gowda2024watt}.

\subsubsection{NVIDIA Jetson TX2}

\paragraph{ImageNet-1K} On TX2, the best model by SAM5 is TinyViT-11M\_Distilled (0.61), which has moderate latency (1008.335 s) and energy (2485.81 J). The reason behind its top performance is the accuracy gains due to knowledge distillation. Compared to its non-distilled version, TinyViT-11M, which has about 20\% lower SAM5 (0.51), TinyViT-11M\_Distilled has higher accuracy (83.8\% vs. 81.0\%) because of knowledge distillation, despite almost identical latency, power, and energy \cite{touvron2021training,wu2022tinyvit}.

Emphasizing energy using SAM1, LeViT\_Conv\_192 and LeViT\_192 have the highest SAM1 (0.26). Both have lowest energies (1067.83 - 1069.49 J) and lowest latencies (500.187 - 605.291 s). The models achieve this via early convolutional downsampling that shortens token sequences and reduces DRAM traffic on TX2’s bandwidth-limited Pascal GPU \cite{graham2021levit,hooker2021moving}. 

\textbf{Takeaway:} On ImageNet-1K, TinyViT-11M\_Distilled is the best when accuracy is the highest priority with SAM5 because of knowledge distillation, whereas the LeViT models lead when energy/time is important with SAM1 due to the hybrid design.

\paragraph{CIFAR-10} 

LeViT\_Conv\_192 is best with SAM5 (1.44) and has the fastest time (295.355 s) and lowest energy (1001.04). It performs better than higher-accuracy models such as ViT\_S (97.1\% vs. 98.6\% ), because its energy is 53\% lower than the baseline ViT\_S (1001.04 vs. 2139.47 J). On TX2 with small 32\(\times\)32 images, LeViT’s early convolutions quickly shrink the token length, so the model moves less data and launches fewer kernels, hence the latency and power are minimized. This is beneficial on edge GPUs \cite{graham2021levit,hooker2021moving}.

Using SAM1, LeViT\_Conv\_192 and LeViT\_192 lead (0.32), with the lowest energies (1001.04 - 1056.09 J) and fastest times (295.355 – 315.243 s). They outperform EfficientViT-B1 (SAM1 = 0.31, 1082.82 J), although EfficientViT-B1 has the lowest power (3338.04 mW), its longer time (324.389 s) raises its energy. LeViT’s early downsampling shortens token sequences \cite{graham2021levit,hooker2021moving}. 

\textbf{Takeaway:} On CIFAR-10, LeViT\_Conv\_192 is the best across metrics on TX2 (best SAM5 and SAM1), which has the lowest latency and energy.

\textbf{NVIDIA Jetson TX2 Takeaway:}
On Jetson TX2, TinyViT-11M\_Distilled leads ImageNet-1K when accuracy is prioritized (SAM5), while LeViT\_Conv\_192 is preferred for energy and time (SAM1). For CIFAR-10, LeViT\_Conv\_192 leads.

\subsubsection{NVIDIA RTX 3050}

\paragraph{ImageNet-1K} 

TinyViT-11M\_Distilled leads (SAM5 = 0.76; 83.8\%, 25.983 s, 19745.50 mW, 512.96 J), about 11\% higher than DeiT-Small\_Distilled (0.68; 82.40\%, 21.338 s, 28727.38 mW, 612.70 J). The key reason is architectural efficiency: TinyViT is hierarchical with stage-wise downsampling and window attention, which shrinks activation sizes and memory traffic; DeiT-Small keeps the vanilla ViT-style global attention at a fixed token count with no hierarchical reduction. As a result, DeiT-Small has more than double the parameters/MACs of TinyViT-11M (22M/4.6G vs. 11M/2.0G), which on RTX 3050 shows up as higher power and energy \cite{wu2022tinyvit,touvron2021training}. Moreover, RTX 3050’s efficient attention kernels (operator fusion and I/O-aware attention) keep transformer throughput high without significant energy consumption \cite{dao2022flashattention}.

TinyViT-11M\_Distilled also tops SAM1 (0.31), outperforming its non-distilled version TinyViT-11M (0.30; 81.00\% 26.542 s, 19324.59 mW, 512.89 J). Despite having the same latency, power and energy, the accuracy gains from distillation makes TinyViT-11M\_Distilled best.

\textbf{Takeaway:} On ImageNet-1K, TinyViT-11M\_Distilled is the best choice (highest SAM5 and SAM1) due to architectural design plus attention-optimized kernels.

\paragraph{CIFAR-10} 

Under SAM5, TinyViT-11M\_Distilled is first (SAM5 = 1.72; 98.6\%, 23.779 s, 21408.38 mW, 509.07 J), even though TinyViT-21M\_Distilled has the highest accuracy (99.3\%). The reason is that relative to TinyViT-11M\_Distilled, TinyViT-21M\_Distilled has more than double parameter count (21.2M vs. 11.0M parameters) and MACs (4.3G vs. 2.0G MACs), which on RTX 3050 leads to much higher power (33.0 W vs. 21.4 W) and energy (838 J vs. 509 J). 

With SAM1, LeViT\_Conv\_192 leads (0.37; 25.893 s, 16049.40 mW,  415.57 J). The metric rewards the lowest energy/time, and LeViT’s fast-inference hybrid design with stage-wise downsampling is the best.

\textbf{Takeaway.} On CIFAR-10, TinyViT-11M\_Distilled wins SAM5 due to lower parameters/MACs, while LeViT\_Conv\_192 leads SAM1 by delivering the lowest energy via its hybrid design.

\textbf{RTX 3050 Takeaway:} On this GPU, TinyViT‑11M\_Distilled is the best choice when accuracy dominates; LeViT\_Conv\_192 is preferred when minimizing energy.

\section{Conclusion and Future Work}
This paper highlights the gap between theoreti-
cal complexity and real-world energy efficiency in ViTs. To address this, we introduced E3P‑ViT, a two‑stage evaluation pipeline combining device‑agnostic screening with device‑related measurement for energy‑aware ViT deployment.  Our extensive benchmarking across TX2 and RTX 3050 on ImageNet‑1K and CIFAR‑10 reveals that energy efficiency is dependent on the specific hardware and task context, and composite theoretical metrics alone can be misleading. Hybrid LeViT‑Conv‑192 reduces energy by up to 53\% on TX2, while TinyViT‑11M\_Distilled dominates on RTX 3050 due to higher accuracy at moderate energy. This research validates that hardware-aware evaluation is essential for sustainable, energy-efficient AI deployment. Future work will include more hardware, datasets and ViT variants.

\bibliographystyle{IEEEtran}
\bibliography{citations}

@inproceedings{gowda2024watt,
  title={Watt for What: Rethinking Deep Learning's Energy-Performance Relationship},
  author={Gowda, Shashank Narayana and Hao, Xinyue and Li, Gen and Sevilla-Lara, Laura},
  booktitle={Computer Vision -- ECCV 2024 Workshops},
  year={2024},
  publisher={Springer}
}

@inproceedings{dosovitskiy2020image,
  title={An image is worth 16x16 words: Transformers for image recognition at scale},
  author={Dosovitskiy, Alexey and Beyer, Lucas and Kolesnikov, Alexander and Weissenborn, Dirk and Zhai, Xiaohua and Unterthiner, Thomas and Dehghani, Mostafa and Minderer, Matthias and Heigold, Georg and Gelly, Sylvain and others},
  booktitle={International Conference on Learning Representations},
  year={2021}
}

@inproceedings{touvron2021training,
  title={Training data-efficient image transformers \& distillation through attention},
  author={Touvron, Hugo and Cord, Matthieu and Douze, Matthijs and Massa, Francisco and Sablayrolles, Alexandre and J{\'e}gou, Herv{\'e}},
  booktitle={International Conference on Machine Learning},
  pages={10347--10357},
  year={2021},
  organization={PMLR}
}

@inproceedings{wu2022tinyvit,
  title={TinyViT: Fast Pretraining Distillation for Small Vision Transformers},
  author={Wu, Kan and Zhang, Jinnian and Peng, Houwen and Liu, Mengchen and Xiao, Bin and Fu, Jianlong and Yuan, Lu},
  booktitle={European Conference on Computer Vision},
  pages={68--85},
  year={2022},
  organization={Springer}
}

@inproceedings{yu2022poolformerv2,
 title = {PoolFormerV2: Efficiently Fusing Local and Global Information for Vision},
 author = {Yu, Weihao and Luo, Mi and Zhou, Pan and Chen, Jiayu and Feng, Jiashi and Yan, Shuicheng},
 booktitle = {Proceedings of the IEEE/CVF Conference on Computer Vision and Pattern Recognition},
 pages = {11634--11644},
 year = {2022}
}

@inproceedings{cai2023efficientvit,
  title={EfficientViT: Lightweight Multi-Scale Attention for High-Resolution Dense Prediction},
  author={Cai, Han and Li, Junyan and Hu, Muyan and Gan, Chuang and Han, Song},
  booktitle={Proceedings of the IEEE/CVF International Conference on Computer Vision},
  pages={17302--17313},
  year={2023}
}

@inproceedings{graham2021levit,
  title={Levit: a vision transformer in convnet's clothing for faster inference},
  author={Graham, Ben and El-Nouby, Alaaeldin and Touvron, Hugo and Stock, Pierre and Joulin, Armand and J{\'e}gou, Herv{\'e} and Douze, Matthijs},
  booktitle={Proceedings of the IEEE/CVF international conference on computer vision},
  pages={12259--12269},
  year={2021}
}

@article{wong2018netscore,
  title={NetScore: Towards universal metrics for large-scale performance analysis of deep neural networks for practical on-device edge usage},
  author={Wong, Alexander and Benzaid, Chafic and Shafiee, Mohammad Javad and Li, Fanjie and Famouri, Mahdi and O'Neil, Brendan},
  journal={arXiv:1806.05781},
  year={2018}
}

@article{hooker2021moving,
  title={The Hardware Lottery},
  author={Hooker, Sara},
  journal={arXiv:2009.06489},
  year={2020}
}

@article{xu2021survey,
  title={A survey on green deep learning},
  author={Xu, Jingjing and Li, Wenhong and Duan, Hang and Xu, Chen},
  journal={arXiv:2111.05193},
  year={2021}
}

@article{schwartz2020greenai,
  title={Green {AI}},
  author={Schwartz, Roy and Dodge, Jesse and Smith, Noah A. and Etzioni, Oren},
  journal={Communications of the ACM},
  volume={63},
  number={12},
  pages={54--63},
  year={2020}
}

@article{tay2020efficient,
  title={Efficient Transformers: A Survey},
  author={Tay, Yi and Dehghani, Mostafa and Bahri, Dara and Metzler, Donald},
  journal={arXiv:2009.06732},
  year={2020}
}

@inproceedings{rao2021dynamicvit,
  title={DynamicViT: Efficient Vision Transformers with Dynamic Token Sparsification},
  author={Rao, Yongming and Zhao, Wenliang and Liu, Benlin and Lu, Jiwen and Zhou, Jie and Hsieh, Cho-Jui},
  booktitle={Advances in Neural Information Processing Systems},
  year={2021}
}

@inproceedings{liang2022evit,
  title={Not All Patches are What You Need: Expediting Vision Transformers via Token Reorganizations},
  author={Liang, Youwei and Ge, Chongjian and Tong, Zhan and Song, Yibing and Wang, Jue and Xie, Pengtao},
  booktitle={International Conference on Learning Representations},
  year={2022}
}

@inproceedings{bolya2023tome,
  title={Token Merging: Your ViT but Faster},
  author={Bolya, Daniel and Fu, Cheng-Yang and Dai, Xiaoliang and Zhang, Peizhao and Feichtenhofer, Christoph and Hoffman, Judy},
  booktitle={International Conference on Learning Representations},
  year={2023}
}

@inproceedings{dao2022flashattention,
  title={FlashAttention: Fast and Memory-Efficient Exact Attention with IO-Awareness},
  author={Dao, Tri and Fu, Daniel Y. and Ermon, Stefano and Rudra, Atri and R{\'e}, Christopher},
  booktitle={Advances in Neural Information Processing Systems},
  year={2022}
}

@inproceedings{chen2018tvm,
  title={TVM: An Automated End-to-End Optimizing Compiler for Deep Learning},
  author={Chen, Tianqi and Moreau, Thierry and Jiang, Ziheng and Zheng, Lianmin and Yan, Eddie and Cowan, Meghan and Shen, Haichen and Wang, Leyuan and Hu, Yuwei and Ceze, Luis and Guestrin, Carlos and Krishnamurthy, Arvind},
  booktitle={Proceedings of the 13th USENIX Symposium on Operating Systems Design and Implementation},
  year={2018}
}

@article{yuan2022ptq4vit,
  title={PTQ4ViT: Post-Training Quantization for Vision Transformers with Twin Uniform Quantization},
  author={Yuan, Zhihang and Xue, Chenhao and Chen, Yiqi and Wu, Qiang and Sun, Guangyu},
  journal={Proceedings of the European Conference on Computer Vision},
  year={2022}
}

@inproceedings{li2022qvit,
  title={Q-ViT: Accurate and Fully Quantized Low-bit Vision Transformer},
  author={Li, Yanjing and Xu, Sheng and Zhang, Baochang and Cao, Xianbin and Gao, Peng and Guo, Guodong},
  booktitle={Advances in Neural Information Processing Systems},
  year={2022}
}

@article{xia2022trtvit,
  title={TRT-ViT: TensorRT-oriented Vision Transformer},
  author={Xia, Xin and Li, Jiashi and Wu, Jie and Wang, Xing and Xiao, Xuefeng and Zheng, Min and Wang, Rui},
  journal={arXiv:2205.09579},
  year={2022}
}

\end{spacing}
\end{document}